# Can ChatGPT capture swearing nuances? Evidence from translating Arabic oaths

Mohammed Q. Shormani, Ibb university, Yemen
shormani@ibbuniv.edu.ye

**Abstract**
This study sets out to answer one major question: Can ChatGPT capture swearing nuances? It presents an empirical study on the ability of ChatGPT to translate Arabic oath expressions into English. 30 Arabic oath expressions were collected from the literature. These 30 oaths were first translated via ChatGPT. We then analyzed and compared these translations to the human translations. Our criteria include gaps left unfulfilled by ChatGPT. Specifically, the gaps involved are: religious gap, cultural gap, both religious and cultural gaps, no gap, using non-oath particles, redundancy and noncapturing of Arabic script diacritics. It concludes that ChatGPT translation of oaths is still much unsatisfactory, unveiling the need of further developments of ChatGPT, and the inclusion of Arabic data on which ChatGPT should be trained including oath expressions, oath nuances, rituals, and practices.

## 1. Introduction

Whether Artificial Intelligence (AI) models like ChatGPT can capture nuances in translation is still a blurring issue. This study investigates whether ChatGPT can accurately capture the linguistic, religious and cultural nuances of oaths in Arabic when translating these expressions into English. Given the complexity of profanity, oaths, and culturally bound expressions, the study explores how well ChatGPT preserves meaning, intensity, and context in its translations, highlighting challenges in machine understanding of what beyond texts.

ChatGPT is one of AI generative models that are capable of producing content in diverse formats, including text, images, music, and beyond, as has been noted by a number of scholars. It has also impacted aspects beyond these, thus revolutionizing the "scientific research, spanning from data processing and hypothesis generation to collaboration and public outreach" (Ray 2023). The revolution taking place in AI sphere is due to several factors, the most important of which is ascribed to the "invasion" of Neural Networking Algorithms (NNA) (see e.g. Peng et al. 2023; Shormani, 2024, a, b & c). These models have also entered translation field, which has revolutionized the translation industry in many and several aspects. There are several AI translation models such as ChatGPT, DeepSeek, Grock 3, and several applications such as Google Translate, Babylon, WordLingo, and numerous other smartphone apps.

ChatGPT is one of these developments; ChatGPT has been developed mainly for translation. However, it has been used for several purposes and in several fields including summarizing, bugging, creating content, completing codes, and automated tagging (Kung et al. 2023; Kenny 2022; Lee 2023; Ray 2023; Siu 2023). Machine translation, or what is known as Computer-Assisted Translation (CAT) tools have been used extensively to translate several types of registers and texts (Kenny 2022; Larroyed 2023).

This article is organized as follows. In section 2, we focus on translation of religious texts, considering what aspects oaths in Arabic involve to make them religious texts. In section 3, we discuss ChatGPT as an AI model, tackling its being a translation application, and how it works in translation. In section 4, we present the study design focusing on the study data, data collection, procedure, and how our data is analyzed. In section 5, the study results are tabularized and then discussed. In section 6, we provide the most important conclusions, shedding light on human postediting as a necessary requirement to fulfill the gaps left untranslated by ChatGPT and providing some recommendations for ChatGPT developers to improve its translation capacity.

## 2. Translating religious and cultural texts

There are several definitions of translations, perhaps as there are translation specialists and theorists. It has been defined one the bases of several aspects and factors. From a linguistic perspective Catford (1965) defines it as rendering the source language text into the target language, maintaining the linguistic features of both languages. He adds that translation should be among the interests of not only linguists or translators, but also other experts in other fields. Additionally, Newmark (1988) defines it as "though not by any means always, [translation] is rendering the meaning of a text into another language in the way that the author intended the text" (: 5). Newmark seems to combine the author's original text and intentions in this definition. Bassnett (2014: 7) argues that translation is "a process of negotiation between texts and between cultures, a process during which all kinds of transactions take place mediated by the figure of the translator". Translation aims for bridging gaps between peoples and nations through several human activities including intercultural communication, mobile marketing, and trade, thus bringing peoples and nations closer, and facilitating a lot of human interactions that were very complicated until very recently (see also Newmark 1981, 1988; Nida 1964; Shormani 2020).

Given that oath expressions are embedded within religion and since religion is inseparable from culture, it follows that oath expressions are religious and cultural texts. Translation of religious texts has been investigated in several studies, holding that this type of text is not easy to translate, because of their religious, cultural and idiosyncratic nature (see also Sirriyya 2009).

To translate culture, one should consider the cultural nuances involved in the text, and then utilize an approach suitable for translating this (cultural) text (Conway 2012). The same applies to religious texts. Equivalence has been proposed to an approach to translating religious texts. According to Nida and Charles (1969), equivalence can be either dynamic or formal. In dynamic equivalence, "[t]he goal ... is to produce in the receptor language the closest natural equivalent of the source-language message, first in meaning and secondly in style" (Nida and Charles 1969: 12). However, the goal of the formal equivalence is "to render the precise contextual meaning of the original words and phrases." The emphasis placed on formal equivalence or correspondence is crucial to Nida for its high significance in successful and appropriate translation. He believes that structure sometimes is more important than wording. In Nida's (1964) approach "the total meaning of the phrase is not signaled by the referential or emotive value of the isolated words *old* and *man*, but a part of the meaning is derived from the constructions itself" (p. 57). Nida postulates that even the ordering of the constituents makes a considerable change in the meaning of the whole construct. Nida emphasized Chomsky's Generative Syntax and Transformational Rules (see Chomsky 1965) as crucial aspects in the translation process, and without which translation may not be accurate. Nida and Charles (1969, p. 24) focus on response and purpose arguing that "there should be a high degree of equivalence of response, or the translation will have failed to accomplish its purpose". Depending on the specific translation purpose, translators sometimes utilize one of these approaches, and some other times, they mix both approaches.

**2.1. Oaths as religious and cultural texts**

Oath expressions are cultural in nature and religious in function. "Oaths are but words, and words but wind" (Samuel Butler, cited in Schlesinger 2008: 15). Conklin (200: 1) defines oath as "a solemn, formal utterance affirming the truth of one's words or expressing one's promise to do something. Oaths can be classified into assertory or promissory; both of these can be either positive or negative". Some scholars view oaths as performative utterances, having the function of strengthening, giving power to, authenticating what we say or what our intercalators say to us (Jay 2009; Qarabesh et al. 2023). Put differently, an oath expression gives to an interlocutor some sort of authentication and makes whatever we say, or said to us authoritative. An oath expression performs a speech act of swearing, swearing on "an entity which is sacred or precious to the swearer, the function of which is perhaps to bear witness to the utterance, or in some way to offer help or lend credence to the oath taker in fulfilling an oath" (Conklin 2005: 1).

Thus, an oath expression is a religious statement. The religiousness in oaths lies in the sacredness of what we swear by. People in different cultures usually swear by entities they glorify, magnify, respect, value, and sometimes love. These entities include Names and/or Attributes of Allah (God) as in "By God!" names of Holy books such as the Holy Qur'an, the Bible, as in "By the Holy Qur'an!", prophets or their qualities as in "By the Prophet Mohammed!". We also swear by entities we value and love such as our children, our lives, as in "On my son's life!", wealth as in "By my wealth!", and more often health as in "On my health!" among other entities. Accordingly, oath statements seem to be expressions of religion. Since religion is embedded within culture, it follows that it is inseparable from culture, and since both are manifested/expressed via language (cf. Newmark, 1988), it is expected that oaths, say, Arabic oaths, are translatable into English, and vice versa. However, the question is: can ChatGPT, an AI model, translate these oath statements?

**2.2. Oaths and human behavior**

According to Qarabesh et al. (2023), our everyday situations, our interlocutors such as friends, relatives, our fellows, colleagues, problems we face form part of our daily life, and communicating with these interlocutors also constitutes part of our cultural wealth. Thus, we find ourselves in these situations, and all other human affairs, relations and communications compelled to confirm, strengthen and/or support the truth of what we say, and/or deny what our interlocutors say to/ about us. These confirmations, supports, strengthenings and denials are delivered by oaths.

An oath unveils the paradox of human language and human behavior; it is not true that a speaker always says the truth. As human behavior, telling the truth seems to constitute a binary system, a logical system. If we consider the possibility of lying an opposite aspect of telling the truth, then telling the truth would be part of our ethical or religious teachings. When we make an oath, we logically make a veridical dimension, simply because no one expects the "other" to always say the truth. It is a feature of human language that a truth is a paradox, and any statement cannot be both true and false, like two faces of the same coin, either this or that, and here lies the concept of "oath". We make oaths to confirm, deny, approve, reject whatever we say or said to us (Qarabesh et al. 2023). Additionally, oath making seems to be universal; all humans make oaths, and since an oath is a piece of language, it is by and large translatable into another language (cf. Brown 1999). If so, the question then is: can ChatGPT capture, in addition to these aspects of human behavior, oath nuances?

**))))))))))**

**3. Machine translation and ChatGPT**

Machine Translation (MT) has been one of the major concerns to involve computer technology in translation endeavor, involving linguistic texts that can be produced and read by computer, and employing "methods for extracting linguistically valuable information from such texts" (Brown et al. 1993: 263). It has developed considerably, specifically recently. It started with employing rule-based mechanism, then statistics, and finally NNA (De Coster et al. 2023), reflecting the stages it has passed through. MT has been made use of in several fields including computer industry, lexigraphy and dictionaries, TV and film industry like subtitling, dubbing, subtitling (Fernández-Costales et al. 2023). With the help of CAT, large amounts of words, passages and even documents can be translated in minutes (Zhang & Torres-Hostench 2022).

However, MT output is often unsatisfactory, specifically concerning translating texts embedded within culture and religion. It can sometimes produce ridiculous and funny translations. Further, MT output is sometimes dangerous, specifically concerning translating medical reports. Such MT output requires human intervention, and depending on the accuracy rates, human translators intervene either in the pre- or in post-editing process (Sahari et al. 2023).

With several and various developments AI industry has undergone, MT has witnessed considerable change. MT's present-day status was a result of such vast and fast developments. CAT tools like Google Translate, Microsoft Translate are just examples of the present-day MT. However, these developments do not stop here, but continue to take place. ChatGPT is a short form of "Chat-Generative Pre-Trained Transformer", a program based on internet, invented by OpenAI in 2018 (Radford et al. 2019), and reported to be one of OpenAI's most influential learning models (Dergaa et al. 2023; Larroyed 2023; Zheng et al. 2023).

There are also thousands of translation apps installed in modern smartphones. ChatGPT, for instance, can translate any amount of data from across languages; it has been fed by huge amounts of internet data through interconnected nodes, what is so-called "recurrent neural networks" Vaswani et al. (2017) or "biological neural networks" (Xu 2022; Kung et al. 2023; Larroyed 2023). Additionally, ChatGPT translation is found to be more correct and appropriate than that of other CAT tools. Jiao et al. (2023) argue that ChatGPT is a good translator, but with GPT-4.

### 3.1. ChatGPT developments

ChatGPT is an example of large language models, a deep learning model that is designed for generating translations much more accurate than other CAT tools such as Google Translate and DeepL (Lee 2023; van Dis et al. 2023). Thus, in this study we explore the ability of ChatGPT to translate an Arabic (Islamic) text, namely oath expressions, into English, examining whether ChatGPT is able to capture religious and cultural nuances involved in these expressions. In fact, religious texts are said to be difficult to translate even by human translators due to the fact that they include religious texts, Allah's Names and Attributes Muslims used to swear by, sacred places like kaʕbah, names of the Holy Qur'an like (مصحف) Muslims also used to swear by, among many others (Sirriyya 2009, cf. also Nida 1964; Nida and Charles 1969; Goodwin 2013).

ChatGPT is an intelligent deep learning language model, based on neural networks. van Dis et al. (2023) argue that ChatGPT is "a machine-learning system that autonomously learns from data and can produce sophisticated and seemingly intelligent writing after training on a massive data set of text" (p. 224). Though the main purpose of ChatGPT is translation, it is also used for several purposes and in several fields including education, medicine, academia, and it can be used in chatting, summarizing, creating content, and automated tagging. In fact, ChatGPT is a revolution in technology industry, conquering several fields. ChatGPT was "initially designed for machine translation... [consisting] of two deep neural networks, namely the encoder and decoder" (Siu 2023: 2). The encoder analyzes the source language word or text and represents internally. However, the decoder "subsequently utilizes to generate target text by predicting each subsequent target word" (ibid). However, though ChatGPT is a powerful language model, it is not able "to browse or perform internet searches" (Kung et al. 2023: 2). Therefore, all responses are generated in situ, based on the abstract relationship between words ("tokens") in the neural network. This contrasts ChatGPT with other chatbots or conversational systems that are permitted to access external sources of information (e.g. performing online searches or accessing databases) in order to provide directed responses to user queries.

The AI developments have launched several conversational AI models e.g. ChatGPT, Gemini Advanced, Claude Sonnet 3, among many others (Ray 2023; Cascella et al. 2023; Sallam et al. 2023). ChatGPT itself has also witnessed several and various developments. Another aspect in which these improvements are manifested could be with regard to storing huge data utilized in the training data, leveraging many tools and algorithms. Today's ChatGPT version is a result of tremendous and continuous work and developments. It is also argued that researching ChatGPT, in particular and AI, in general, and how to develop it will be in the forefront for decades to come (Lee 2023; Ray 2023). It is also true that computer-and-smartphone-based technologies will also be continuing for the foreseeable future (see Rodgers et al. 2023). ChatGPT has witnessed several developments, and these developments can be clearly seen in its models: ChatGPT-1, GPT-2, GPT-3, GPT-4.

Further, as reported by Ray (2023: 122; see also Cascella et al. 2023; Sallam et al. 2023), there are several developments ChatGPT has undergone, involving a series of milestones and improvements, including.

1. The introduction of the Transformer architecture, which enabled the creation of highly efficient and scalable language models.
2. The development and release of the GPT series, which demonstrated the potential of AI language models in various applications, including text generation, translation, and summarization.
3. The release of ChatGPT, which built upon the successes of its predecessors while incorporating improvements in accuracy, context understanding, and versatility.

ChatGPT has entered many fields, and more importantly the academic sphere, and students and researchers' life, in particular (Cotton et al. 2023; Dergaa et al. 2023; Lee 2023). Dergaa et al. (2023), for instance, argue that ChatGPT is able "to create well-written student essays, summarize research papers, answer questions well enough to pass medical exams, and generate helpful computer codes". It has several consequences far-reaching "for science and society" (van Dis et al. 2023: 224).

## 4. Methodology

### 4.1. Data collection

The collection of the data used in this study started with selecting 30 (Islamic) Arabic oath expressions from Qarabesh et al. (2023). The purpose of selecting these expressions from Qarabesh et al. (2023) is that they have ready human translation by these authors so that the results of ChatGPT translation could be easily and reliably compared and contrasted with human translation by these authors. Comparing ChatGPT translation to human translation would reveal the gaps left untranslated by ChatGPT, types of these gaps and the degree of deviation.

### 4.2. Procedure

This study is empirical in nature; it took place in three stages. Stage one was the collection of the 30 Arabic oath expressions. Stage two was devoted to translating these expressions into English via ChatGPT. Stage three concerned analyzing, comparing and contrasting ChatGPT translations with human translation already available from Qarabesh et al. (2023). We analyze the data in terms of no gap left, religious gap, cultural gap, cultural and religious gaps, redundancy, noncapturing of Arabic diacritics and using non-oath particles. Figure 1 displays a sample of ChatGPT translation of the Arabic oath expressions.

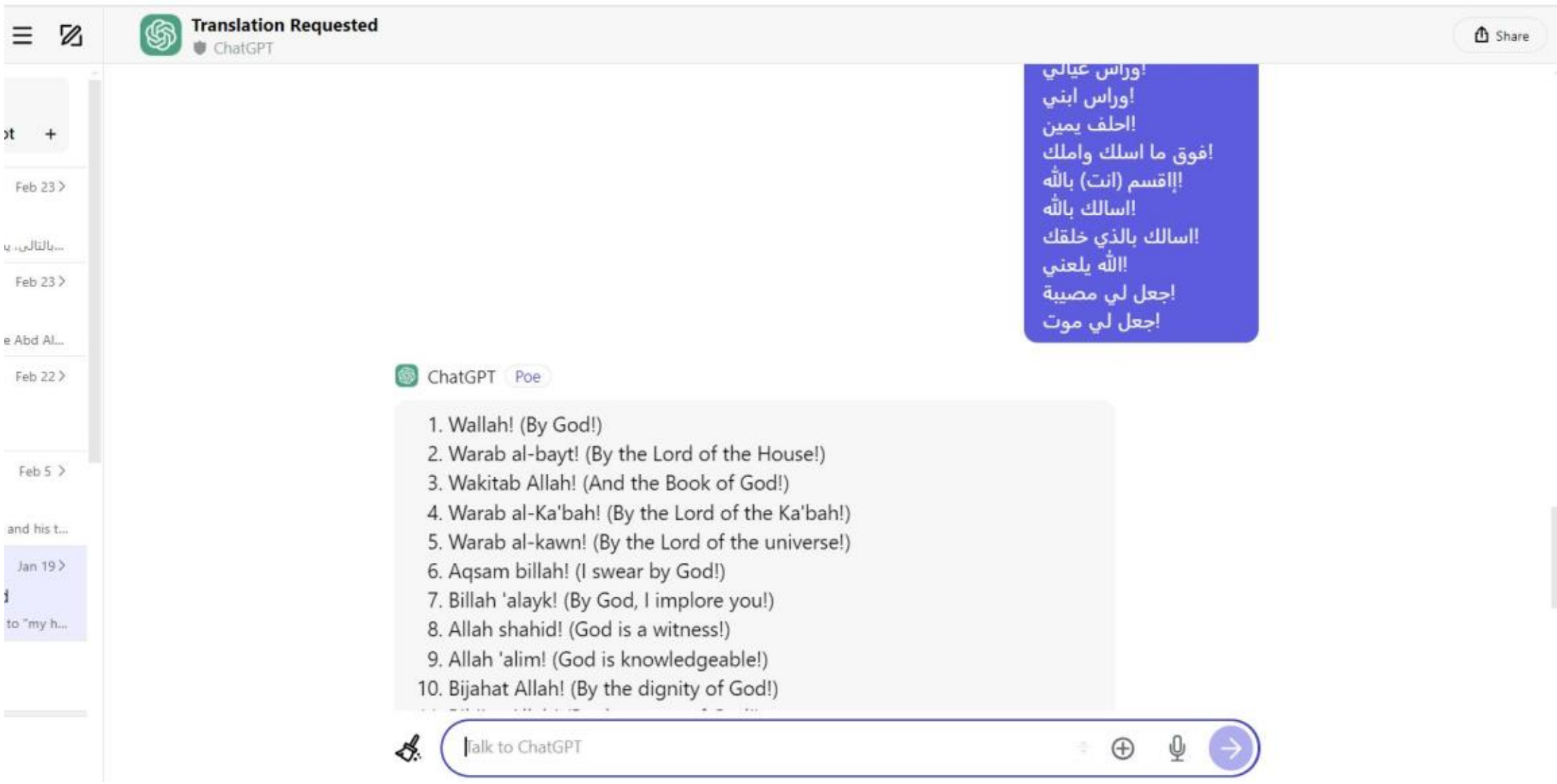

**Figure 1: A sample of ChatGPT translation**

For the full list of ChatGPT translation of the 30 oaths involved**,** see **Appendix A**.

### 4.3. Method of analysis

Given the analysis and comparison nature of this study, the method of analyzing the data is an amalgamated method based on the analytical and comparative approaches.

## 5. Results and discussion

We first tabulate the results and then discuss them. Section 5.1. presents the results and section 5.2. discusses these results.

### 5.1. Results

Table 1 presents the study results in terms of oath expression, transliteration, ChatGPT translation and human translation.

**Table 1: Arabic, Transliteration, ChatGPT translation, Human Translation**

| Arabic text | Transliteration | ChatGPT translation | Human translation |
|---|---|---|---|
| والله! | wa-llah! | By God! | By God! |
| ورب البيت! | wa-rrab l-bayt! | By the Lord of the House! | By the Lord of kaʕbah! |
| وكتاب الله! | wa-kitab ʔallah!! | And the Book of God! | (I swear) by the Qur'an! |
| ورب الكعبة! | wa-rrab l-kaʕbah! | By the Lord of the Kaʕbah! | By the Lord of kaʕbah! |
| ورب الكون! | wa-rrab al-kawn! | By the Lord of the universe! | By the Lord of Universe! |
| اقسم بالله! | ʔ𝑖qsam billah! | I swear by God! | I swear by God! |
| بالله عليك! | billah 'ʕalayk! | By God, I implore you! | Swear me! |
| الله شاهد! | ʔallah! šahid! | God is a witness! | God is witness |
| الله عالم ! | ʔallah! ʕalim! | God is knowledgeable! | God knows! |
| بجاه الله! | bi-jah ʔallah!! | By the dignity of God! | For God's sake! |
| بحجر الله! | bi-ħajr ʔallah!! | By the stone of God! | For God's sake! |
| فوق المصحف! | fawq l-musħaf! | Above the Quran! | By the Qur'an! |
| فوق الختمة! | fawq al-khatmah! | Above the rosary! | By the Qur'an! |
| عدمني عيوني! | ʕaddamni ʕiūni! | My eyes fail me! | God blind me! |
| عليا الطلاق! | ʕaliya al-talaq! | "Aliya divorces me" | By the Sacredness of Marriage! |
| والذي خقلك! | wa- llaði xalaqak! | By the One who created you! | By God! |
| إقسم بالذي خلقك! | ʔiqsam bi-llaði halaqak! | I swear by the One who molded you! | Swear by God! |
| عدمني قرآني! | ʕaddamni Qur'ani! | My Qur'an eludes me! | By (my memorization of) the Qur'an! |
| فوق ركبتي! | fawq rukbati! | Above my knee! | By my health! |
| وراس عيالي! | wa-ras ʕayyali! | By the heads of my children! | On the lives of my children! |
| وراس ابني! | wa-ras ʔibni! | By the head of my son! | On the life of my son! |
| احلف يمين! | ʔiħlaf yamin! | I swear by my right hand! | Swear me! |
| فوق ما اسلك واملك! | fawq ma-asaluk wa-amlak! | Above what I walk and possess! | By my wealth! |
| إقسم بالله! | aqsam bi'llah | I swear by God! | Swear me! |
| اسالك بالله! | ʔasʔaluk billah! | I ask you by God! | Swear me by God! |
| اسالك بالذي خلقك! | ʔasʔaluk bi-llaði xalaqik | I ask you by the One who created you! | Swear me by God! |
| الله يلعني! | ʔallah! yalʕani! | God curse me! | God curse me! |
| جعل لي مصيبة! | Ja'al li musībah! | May a calamity befall me! | God curse me! |
| جعل لي موت! | jiʕl li maut! | May death befall me! | I hope to die! |
| وشرفي! | wa-šarafi! | And my honor! | By my honor! |

**5.2. Discussion**

Scrutinizing Table 1, we will analyze ChatGPT translation in terms of the following types:

**No gap**

This section presents the analysis of the oath expressions where ChatGPT translation meets human translation. For example, والله!'" was translated correctly similarly by both ChatGPT and human as" By God!" . Another example can be "ورب الكعبة!" which was translated by both ChatGPT and human as "By the Lord of (the) kaʕbah!" (we will not consider transliteration errors). Further examples of this phenomenon includes translating the Arabic oaths "ورب الكون!", "اقسم بالله!", "الله يلعني!", among others.

The reason of ChatGPT's success in translating these expressions could be attributed to the fact that these expressions are simple and used thoroughly by Muslims worldwide as in the case of "اقسم بالله". Another reason might be that these expressions are culturally well-known, specifically at the age of technology and globalization. This means that ChatGPT has perhaps been trained on these expressions, given the massive data it has been fed with up to 2021.

**Religious gap**

In this category, we examine those oaths where ChatGPT left a religious gap in its translation. A religious gap that is left untranslated/unaccounted for by ChatGPT seems to prevail throughout most of the 30 oaths. Thus, this category can be exemplified by cases such as translating "ورب البيت!" which was translated by ChatGPT as "By the Lord of the House!", while human translates it as "By the lord of kaʕbah". In this particular example, ChatGPT seems to favour or consider the simple sense, or only denotation of the word "house". It is well known that when a Muslim says "رب البيت" or swears " ورب البيت!", it is religiously understood that what is meant by "البيت" 'house' is not any ordinary house, but rather "kaʕbah!". Another example of this phenomenon is translating "بالله عليك!", which simply means 'Swear me!'. However, ChatGPT translates it as "By God, I implore you!" where there is a religious gap left unfulfilled. In fact, there is nothing like "imploring" or appealing here. The situation in which this oath is said could be that the speaker says something, and the hearer did not believe that as a surprise, and so the hearer wanted the speaker to affirm this.

Further examples of this category include "!بجاه الله", which ChatGPT translates as "By the dignity of God!". From a first glance, it seems that the word "dignity" fulfils the intended meaning, but it does not do so. The sense in which a word can be used here is "greatness" (of God) and not "dignity", thus again ChatGPT left a gap here. This oath can be translated into English as "For God's sake!'" which we find in human translation. The most serious religious gap left by ChatGPT is regarding translating the oath "!بحجر الله". ChatGPT translated it as "By the stone of God!" translating the word 'حجر" as "stone" which is not only inappropriate, but completely incorrect. The word "حجر" here does not mean stone, but rather "sake". Here, ChatGPT seems not capture the diacritic use of Arabic writing system as "حَجَرْ" is different from "حَجْرْ".

## Cultural gap

As for the cultural gap left untranslated by ChatGPT, it is clear that most of the ChatGPT translations involve it. Let's take translating the oath "!جعل لي مصيبة" as an example which was translated by ChatGPT as "May a calamity befall me!". In Yemeni Arabic, and specifically in this context it means "curse" "مصيبة", and the meaning of the oath as a whole is "God curse me!" which was done by human translation. Another example of this phenomenon could be translating the oath "!فوق الختمة". This oath was translated by ChatGPT as "Above the rosary!", which is not only culturally odd, but also meaningless. The meaninglessness of such a translation lies in translating "الختمة" as "rosary" whose meaning is so far from "الختمة". "الختمة" means "Qur'an", and "By the Qur'an!" is the most appropriate translation, as human translates it.

Another cultural gap left by ChatGPT untranslated concerns translating the oath " احلف يمين!" which literally means 'swear me an oath'. And in this oath context, it simply means "Swear me!", which, in turn, is an English oath. However, this oath was translated by ChatGPT as "I swear by my right hand!" which is neither accurate nor appropriate, where "يمين" 'oath' was translated as 'right hand'. "يمين" in Yemeni culture means 'oath'. Human translation got it right, where the oath was translated as "Swear me!". Another cultural gap left by ChatGPT can be noticed in translating the oath " فوق ما اسلك واملك", whose ChatGPT translation is "Above what I walk and possess!". The unnaturalness of this translation lies in several aspects: i) the preposition "above" is not a swearing particle/word in English, and ChatGPT seems to have translated it literally, ii) the Yemeni word "اسلك" in this context means "what I owe but not under my control", and ChatGPT translated it as "walk", which is even far from the Arabic term "مسلك meaning "path". iii) though the word "املك" means "possess", ChatGPT fails to capture the cultural (and even religious) nuances here, and iv) both Yemeni terms "اسلك واملك" mean "wealth", and so human translators capture this meaning and translate the whole oath as "By my wealth!".

## Cultural and religious gaps

In this category, we address both cultural and religious gaps that are simultaneously left unfulfilled by ChatGPT. To start with, the oath "!جعل لي موت" was translated by ChatGPT as “May death befall me!”. This translation is simply misleading for several reasons: i) ‘death’ cannot be sworn by, both in English and Arabic cultures. Religiously, death cannot “befall” someone, but rather takes someone and moves him/her from a life to another life, i.e. meta-life, a life that is beyond human conception, as believed by Muslims, and ii) the oath “جعل لي موت!” in Yemeni Arabic, and perhaps across other Arabic varieties means “if I lie, I hope to die”.

As noted by Qarabesh et al. (2023: 13), the oath “I hope to die!” reflects the swearer’s hope to die, which is “similar to the oath said by Harry, in the movie Home Alone 2”. The situation of this oath, they argue, is when the two robbers wanted to harm Kevin “a ten-year old child” (ibid: 11). And iii) hoping to die is also a feature of both cultures, viz., Arabic and English, and also in both religions, i.e. Islam and Christianity. Put simply, in Arabic culture, for instance, when one is overindebted, one hopes to die to just be relieved of one’s debt. It is also there when one feels that life becomes difficult and unbearable.

Another example of both cultural and religious gaps can be observed in ChatGPT translation of “عدمني قرآني!”, which was translated by ChatGPT as “My Quran eludes me!”. This translation is nonsense; it neither transmits the religious nor the cultural intended meaning. In Muslim culture, when one memorizes the Holy Qur’an, one is considered to be superior to others who do not do so. Thus, in Islam we cannot say the Holy Qur’an “eludes” a memorizer. In Islam also, we swear by losing the Holy Qur’an because memorizing it is something extraordinary, and the real entity sworn by here is God Who is the only One Who takes away Qur’an from a memorizer. Here, we think Qarabesh et al.’s translation was not very much successful. They translate it as “By (my memorization of) the Qur’an!”, the most appropriate and accurate translation would be “By Who can take my Qur’an!”

The last example of cultural and religious gaps can be found in translating the oath “عليا الطلاق!”, which was translated by ChatGPT as “Aliya divorces me”, which, in turn, is neither culturally nor religiously appropriate. ChatGPT translates “ عليا“ (literally ‘on me’) as “Aliya”, a name of a woman. Culturally, “Aliya” cannot divorce, because in Islam only man can divorce, but not woman. Culturally also, in English it is not possible to swear by “divorce”. Additionally, from a religious perspective, however, not only in Islam or Christianity, but perhaps in all religions, marriage life is sacred, and so translating this oath as “By the Sacredness of Marriage!” is the most appropriate method, both religiously and culturally (Qarabesh et al., 2023). It transmits the intended meaning of the source language, i.e. Arabic, and at the same time English natives understand it clearly.

**Redundancy**

As can be seen in Table 1, there are also several examples of redundancy in ChatGPT translation. Regardless of being a correct, appropriate, accurate translation or not, ChatGPT translation seems to be redundant as a result of w-f-w method of translation. For example, it translates the oath “اسالك بالله!” as “I ask you by God!”, whereas human translates it as "Swear me by God!". Another example of this phenomenon is "! اسالك بالذي خلقك" which was translated by ChatGPT as "I ask you by the One who created you!", which is redundant compared to human translation, which again is "Swear me by God!". The verb "اسال" here means "swear", but ChatGPT translates it as "ask", which is very far from the context of the oath. A further example of redundancy could be related to the translation of the oath "إقسم بالذي خلقك!" translated by ChatGPT as "I swear by the One who molded you! and "جعل لي مصيبة!", which was translated as "May a calamity befall me!". Additionally, this ChatGPT translation is "translation by addition" due to the fact that if we add, the text is altered inappropriately.

**Noncapturing of Arabic diacritics**

Arabic language script has what is known as diacritics. These are (-َ, -ِ, -ْ), the first two of which can be represented by the English short vowel sounds /a/, /i/ and the third is just no sound or /φ/. ChatGPT could not capture these elements in its translation. The first example could be translating the oath "بحجر الله!". As noted above, ChatGPT translated it as "the word "حجر" as "stone" which is a result of not capturing the Arabic diacritics. The word "حجر" here does not mean stone, but rather "sake". It is true that these diacritics are not written on the word, but it is easy for an Arab to notice these things depending on the context in which the word is used. What is meant here is not "حَجَرْ", but "حَجْرْ", which means in an oath for God's sake. Another example of this phenomenon is regarding the translation of the word "إقسم", in two oaths: "! إقسم بالذي خلقك" and "إقسم بالله!". In these two oath contexts, the speaker is asking the hearer (his/her interlocutor) to confirm what he/she is saying by making an oath asking him/her to "swear". ChatGPT takes it as أقسم where "ء" or hamza is above "ا" or aliph. In Arabic when hamza is on aliph it is different from when it is below aliph. In Arabic writing system, there is a difference between " أقسم" and " إقسم". In the former, it is the speaker who swears, and in the latter, it is the hearer who should swear.

**Using non-oath particles**

According to Qarabesh et al. (2023), Arabic swearing particles are: *wa*, *bi*-, *fauq* (*and*, *by*, and *on*, respectively) and the classic Arabic particle *ta*-. They also argue that English swearing particles/words are *by* and *on.* However, in ChatGPT translation we found that several non-oath particles/words have been used, including "above" and "and" as swearing particles/words. "and" was used in relation to the oath "وكتاب الله!" in "And the Book of God!" and "above" was used four times. For example, in translating the oath "فوق المصحف!" "above" was used "Above the Quran!". Another example is regarding translating the oath "فوق الختمة!" as "Above the rosary!". Though the verb or auxiliary "may" is not a swearing particle/word in English, it has been used two times by ChatGPT.

## 6. Conclusion

AI such as CAT tools including Google Translate have been extensively used in translation and nontranslation industries. As for translation, different types of texts have been translated from one language into another language such as translating technical, nonliteral, noncultural and nonreligious topics. However, considering religious and cultural words, expressions and texts such as oath expressions, it is clear that even ChatGPT is still in need of more developments and improvements to translate and capture religious and cultural nuances. Put differently, although ChatGPT has succeeded in several aspects including writing summaries, computer codes, researchers' literature reviews, talks and presentations, statistical analyses, among others (van Dis et al. 2023), it seems that it is not very much successful in translation, specifically regarding translating religious and cultural expressions. This means that ChatGPT needs more developments and improvements.

Thus, based on the results of this study, and the gaps left untranslated by ChatGPT when translating Arabic oath expressions, we can conclude that ChatGPT translation is not that much satisfactory. There are a number of flaws in ChatGPT translation, some are very serious, some are serious to some extent and some are not. At least for the time being, we recommend that these gaps left unfulfilled by ChatGPT need translators to intervene in the post-editing process. Postediting is simply defined as "the correction of raw machine-translated output by a human translator according to specific guidelines and quality criteria" (O'Brien 2011: 197-198). In our study, it is a following-up translation process carried out by human translators to improve ChatGPT translation (cf. de Almeida & O'Brien 2010; Groves & Schmidtke 2009; Krings 2001; O'Brien 2007, 2011).

This study provides substantial insights that human postediting is still necessary, and without it ChatGPT translation, specifically concerning religious and cultural aspects would be full of inaccuracies and inappropriateness. Thus, it is proposed that human translator can first use ChatGPT to produce "first draft translation" which must be followed by postediting. This postediting should cover all aspects left unfulfilled/untranslated by ChatGPT. The human postediting could involve areas which ChatGPT is not trained on. These include Islamic religious and cultural texts, in general involving religious beliefs, nuances, sacred terms, oaths, names of the Holy Qur'an, Names and Attributes of Allah. This is the way human role could be integrated after ChatGPT translation.

**Conflict of interest:**
There is no conflict of interest to declare.

**Funding:**
The author received no funding for this research.

**Appendix A: ChatGPT translation of the 30 oath expressions**

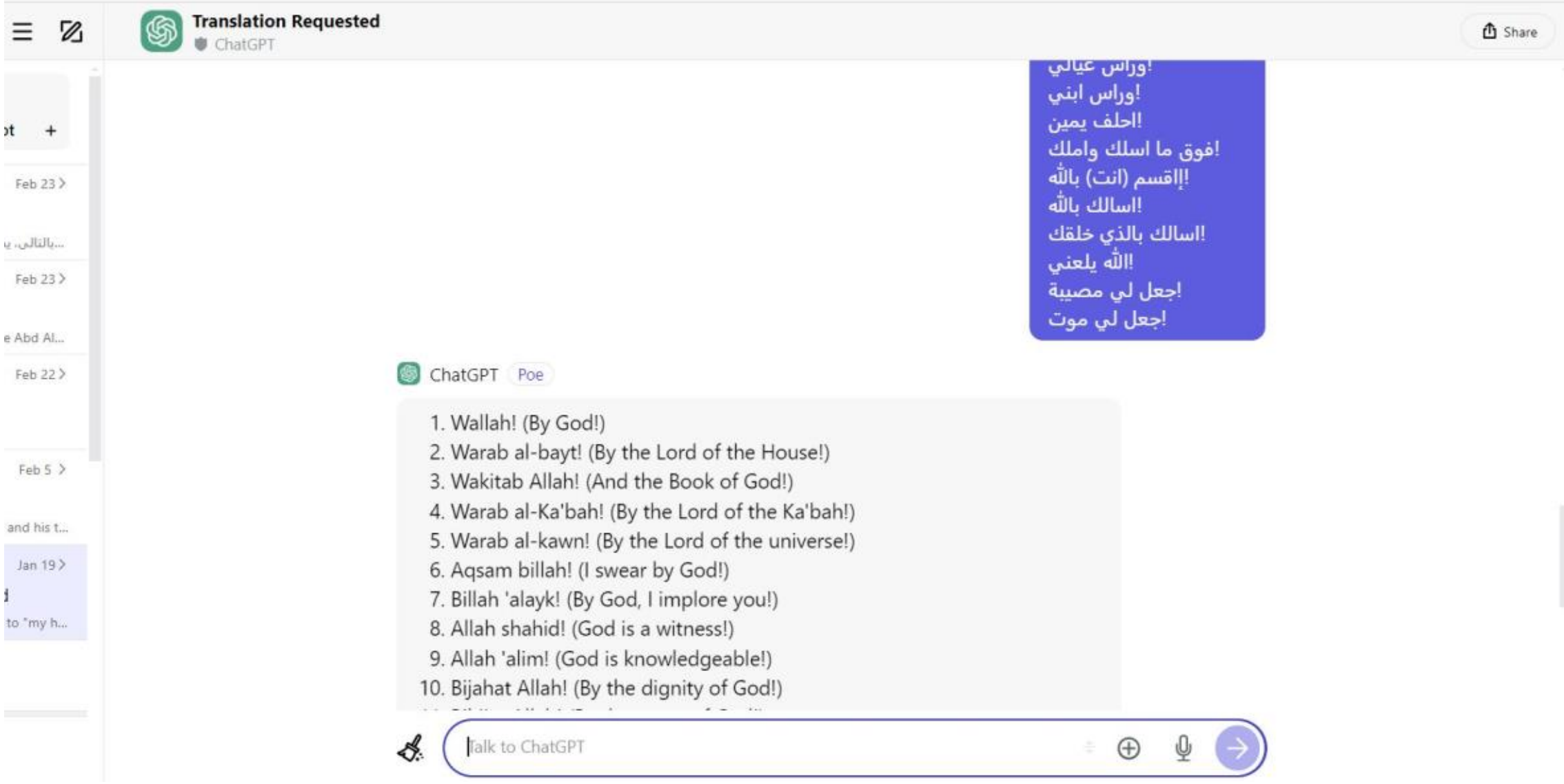

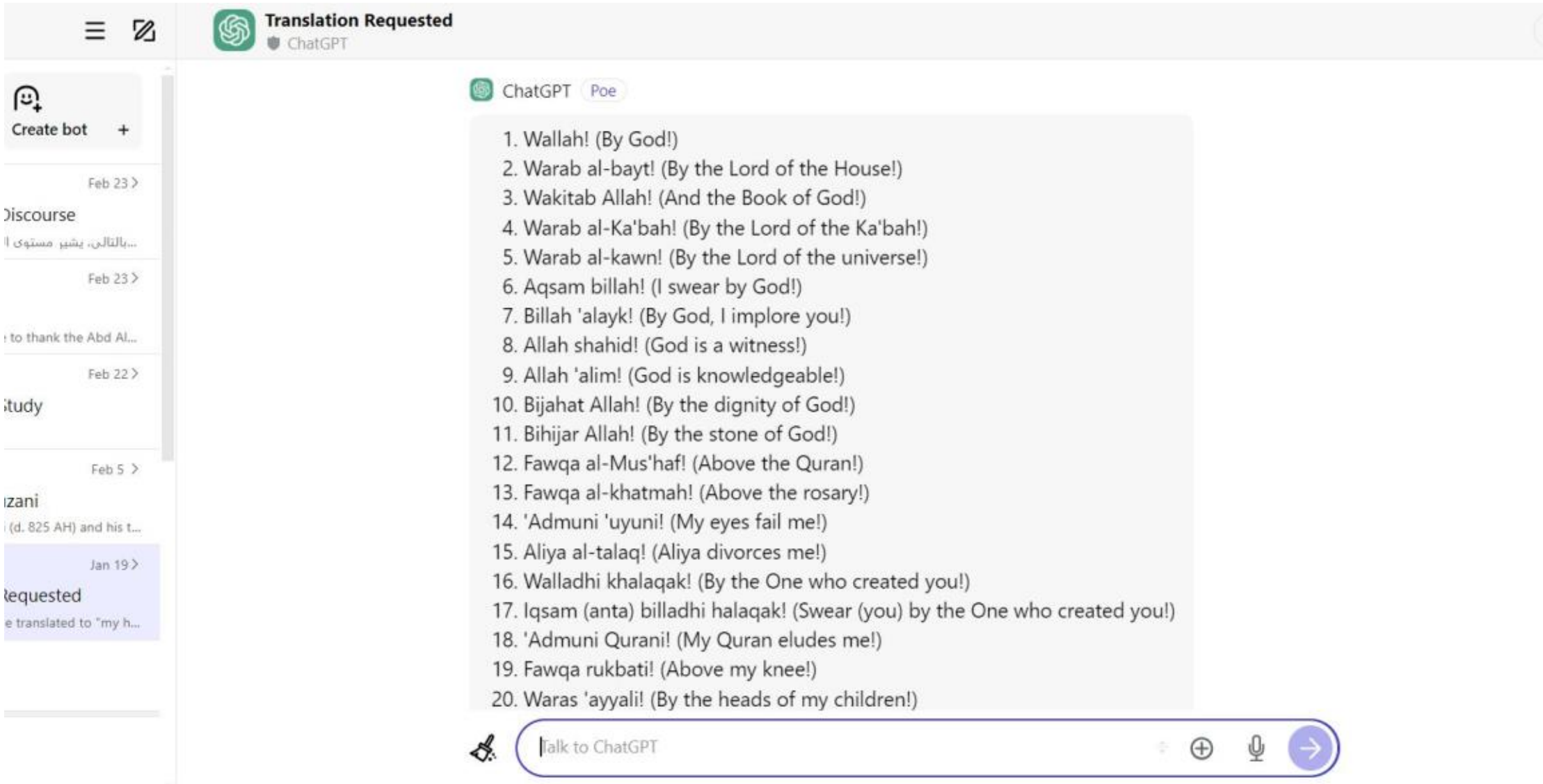

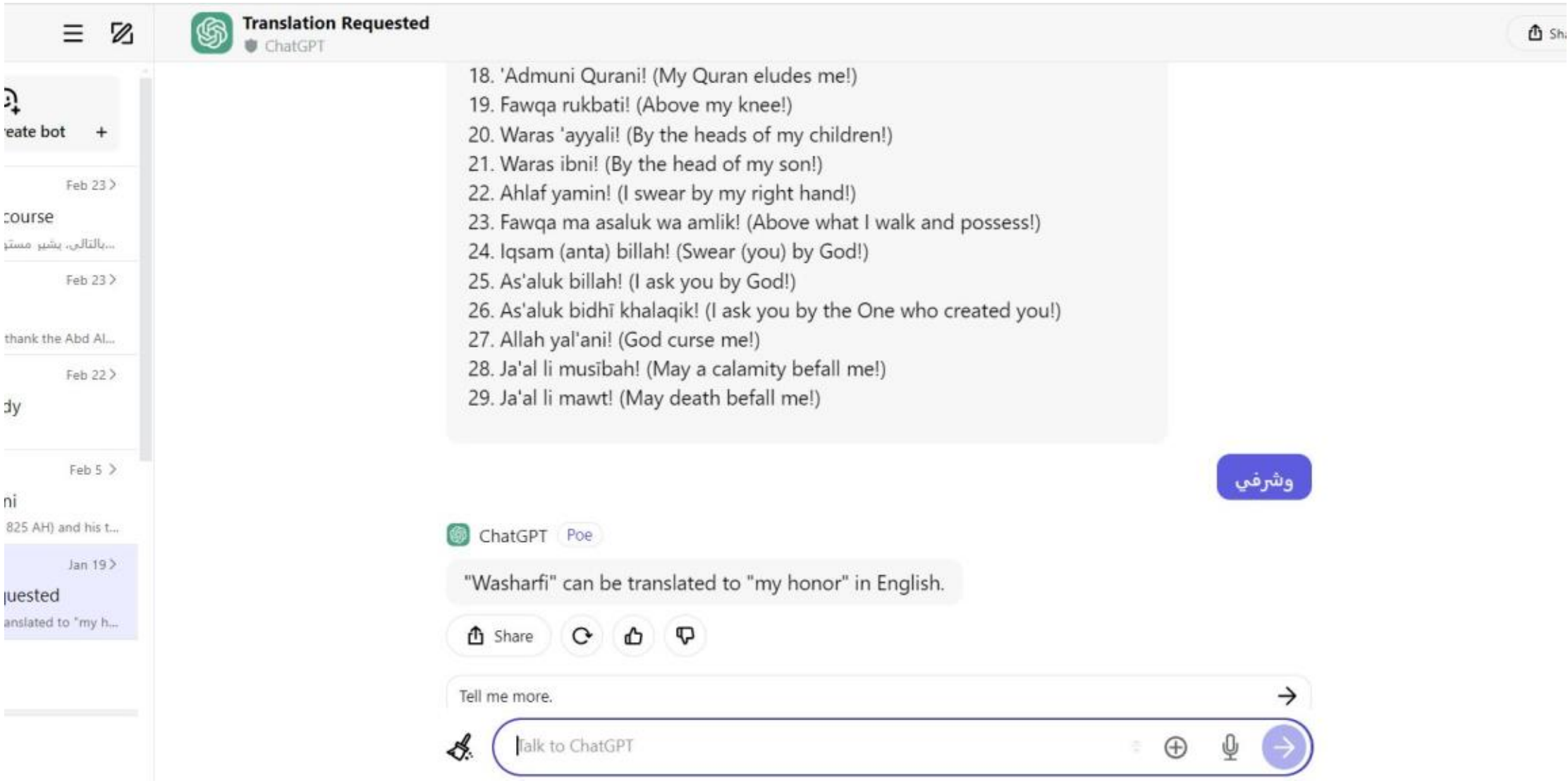
Translation Requested
ChatGPT
18. 'Admuni Qurani! (My Quran eludes me!)
19. Fawqa rukbati! (Above my knee!)
20. Waras 'ayyali! (By the heads of my children!)
21. Waras ibni! (By the head of my son!)
22. Ahlaf yamin! (I swear by my right hand!)
23. Fawqa ma asaluk wa amlik! (Above what I walk and possess!)
24. Iqsam (anta) billah! (Swear (you) by God!)
25. As'aluk billah! (I ask you by God!)
26. As'aluk bidhī khalaqik! (I ask you by the One who created you!)
27. Allah yal'ani! (God curse me!)
28. Ja'al li musībah! (May a calamity befall me!)
29. Ja'al li mawt! (May death befall me!)
وشرفي
ChatGPT Poe
"Washarfi" can be translated to "my honor" in English.
Share
Tell me more.
Talk to ChatGPT